\pdfoutput=1
\documentclass[11pt]{article}

\usepackage{mypackage}

\usepackage{acl}

\usepackage{times}
\usepackage{latexsym}

\usepackage[T1]{fontenc}

\usepackage[utf8]{inputenc}

\usepackage{microtype}

\title{Improving Grammar-based Sequence-to-Sequence Modeling with Decomposition and Constraints}

\author{Chao Lou, Kewei Tu\thanks{\; Corresponding Author}\\
  School of Information Science and Technology, ShanghaiTech University \\
  Shanghai Engineering Research Center of Intelligent Vision and Imaging\\ 
    {\tt \{louchao,tukw\}@shanghaitech.edu.cn}\\
 }

\begin{document}
\maketitle
\begin{abstract}
Neural QCFG is a grammar-based \ac{seq2seq} model with strong inductive biases on hierarchical structures. It excels in interpretability and generalization but suffers from expensive inference. In this paper, we study two low-rank variants of Neural QCFG for faster inference with different trade-offs between efficiency and expressiveness. Furthermore, 
utilizing the symbolic interface provided by the grammar, we introduce two soft constraints over tree hierarchy and source coverage. We experiment with various datasets and find that our models outperform vanilla Neural QCFG in most settings.
\end{abstract}

\section{Introduction}

Standard neural \ac{seq2seq} models are versatile and broadly applicable due to its approach of factoring the output distribution into distributions over the next words based on previously generated words and the input~\cite{sutskever2014sequence,gehring2017convolutional,devlin-etal-2019-bert}. 
Despite showing promise in approximating complex output distributions, these models often fail when it comes to diagnostic tasks involving compositional generalization~\cite{lake2018generalization,bahdanau2018systematic,loula-etal-2018-rearranging}, possibly attributed to a lack of inductive biases for the hierarchical structures of sequences (e.g., syntactic structures), leading to models overfitting to surface clues.

In contrast to neural \ac{seq2seq} models, traditional grammar-based models incorporate strong inductive biases to hierarchical structures but suffer from low coverage and the hardness of scaling up~\cite{wong-mooney-2006-learning,bos-2008-wide}. 
To benefit from both of these approaches, blending traditional methods and neural networks has been studied~\cite{herzig-berant-2021-span,shaw-etal-2021-compositional,wang2021structured,wang2022hierarchical}. 
In particular, \citet{NEURIPS2021_dd17e652} proposes Neural QCFG for seq2seq learning with a \ac{qcfg}~\cite{smith-eisner-2006-quasi} that is parameterized by neural networks. 
The symbolic nature of Neural QCFG makes it interpretable and easy to impose constraints for stronger inductive bias, which leads to improvements in empirical experiments.
However, all these advantages come at the cost of high time complexity and memory requirement, meaning that the model and data size is restricted, which leads to a decrease in text generation performance and limited application scenarios.

In this work, we first study low-rank variants of Neural QCFG for faster inference and lower memory footprint based on tensor rank decomposition~\cite{https://doi.org/10.48550/arxiv.1711.10781}, which is inspired by recent work on low-rank structured models~\cite{cohen-etal-2013-approximate,chiu2021low,yang-etal-2021-pcfgs,yang-etal-2022-dynamic}. 
These variants allow us to use more symbols in Neural QCFG, which has been shown to be beneficial for structured latent variable models~\cite{pmlr-v119-buhai20a,chiu-rush-2020-scaling,yang-etal-2021-pcfgs,yang-etal-2022-dynamic}.
Specifically, we study two low-rank variants with different trade-off between computation cost and ranges of allowed constraints: the \underline{e}fficient model (E model), following the decomposition method in TN-PCFG~\cite{yang-etal-2021-pcfgs}, and the ex\underline{p}ressive model (P model), newly introduced in this paper. Furthermore, we propose two new constraints for Neural QCFG, including a soft version of the tree hierarchy constraint used by vanilla Neural QCFG, and a coverage constraint which biases models in favour of translating all source tree nodes\footnote{Similar topics are discussed in the machine translation literature \cite[among others]{tu-etal-2016-modeling,li-etal-2018-simple}.}.
We conduct experiments on three datasets and our models outperform vanilla Neural QCFG in most settings. Our code is available at \href{https://github.com/LouChao98/seq2seq_with_qcfg}{https://github.com/LouChao98/seq2seq\_with\_qcfg}.

\section{Preliminary: Neural QCFG}
\label{sec:preliminary}

Let $s_1,s_2$ be the source and target sequences, and $t_1,t_2$ be the corresponding constituency parse trees (i.e., sets of labeled spans). Following previous work~\cite{smith-eisner-2006-quasi,NEURIPS2021_dd17e652}, we consider \ac{qcfg} in Chomsky normal form~\cite[CNF;][]{CHOMSKY1959137} with restricted alignments,
which can be denoted as a tuple $G[t_1]=(S, \NT, \PT, \Term, \Rule{t_1}, \theta)$, where $S$ is the start symbol, $\NT/\PT/\Term$ are the sets of nonterminals/preterminals/terminals respectively, $\Rule{t_1}$ is the set of grammar rules in three forms:
\begin{align*}
  & S \rightarrow A[\alpha_i] \quad\text{where }A\in \NT,\ \alpha_i\in t_1,                   \\
  & A[\alpha_i] \rightarrow B[\alpha_j] C[\alpha_k]  \text{ where } \\
  & \qquad A \in \NT,\ B,C\in\NT\cup\PT,\ \alpha_i, \alpha_j, \alpha_k\in t_1,           \\
  & D[\alpha_i] \rightarrow w  \quad\text{where }A\in \PT,\ \alpha_i\in t_1,\ w\in\Sigma,
\end{align*}
and $\theta$ parameterizes rule probablities $p_\theta(r)$ for each $r\in \Rule{t_1}$.

Recently, \citet{NEURIPS2021_dd17e652} proposes Neural QCFG for seq2seq learning. He uses a source-side parser to model $p(t_1|s_1)$ and a QCFG to model $p(t_2|t_1)$. The log marginal likelihood of the target sequence $s_2$ is defined as follows:
\begin{align*}
  &\log p_{\theta,\phi}(s_2 | s_1)  \\
  &=\quad \log \sum_{t_1\in\T(s_1)} p_\theta(s_2| t_1) p_\phi(t_1|s_1) \\
  &=\quad \log \sum_{t_1\in\T(s_1)}\sum_{t_2\in\T(s_2)} p_\theta(t_2| t_1) p_\phi(t_1|s_1),
\end{align*}
where $\T(\cdot)$ denotes the set of possible parse trees for a sequence and $\theta,\phi$ are parameters. Due to the difficulty of marginalizing out $t_1$ and $t_2$ simultaneously, \citet{NEURIPS2021_dd17e652} resorts to maximizing the lower bound on the log marginal likelihood,
\begin{align*}
  \log p_{\theta,\phi}(s_2 | s_1) \ge \mathbb{E}_{t_1\sim p_\phi(t_1|s_1)}\left[\log p_\theta(s_2|t_1)\right].
\end{align*}


\section{Low-rank Models}

\begin{figure}[tb]
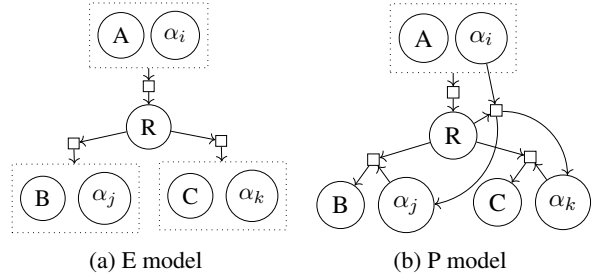

  \centering
  \begin{subfigure}[b]{0.48\columnwidth}
    \centering
    \resizebox{\columnwidth}{!}{
    \tikz {
      \node [circle,draw] (A)              {A};
      \node [circle,draw] (i) [right=0.1cm of A] {$\alpha_i$};
      \node [draw,dotted,fit=(A) (i)] (Ai) {};
      \node [draw] (AiR) [below=0.2cm of Ai,inner sep=0pt,minimum size=5pt] {};
      \node [circle,draw] (R) [below=0.2cm of AiR] {R};
      \node [circle,draw] (j) [below left=0.6cm and 0.15cm of R] {$\alpha_j$};
      \node [circle,draw] (B) [left=0.2cm of j] {B};
      \node [draw] (RjB) [above=0.8cm of $(B)!0.5!(j)$,inner sep=0pt,minimum size=5pt] {};
      \node [circle,draw] (C) [below right=0.6cm and 0.15cm of R] {C};
      \node [circle,draw] (k) [right=0.2cm of C] {$\alpha_k$};
      \node [draw] (RkC) [above=0.8cm of $(C)!0.5!(k)$,inner sep=0pt,minimum size=5pt] {};
    \node [draw,dotted,fit=(B) (j)] (Bj) {};
    \node [draw,dotted,fit=(C) (k)] (Ck) {};
    \draw [->]
    (AiR) edge (R)
    (RjB) edge (Bj)
    (RkC) edge (Ck)
    (Ai) edge (AiR)
    (R) edge (RjB)
    (R) edge (RkC);
  }}
    \caption{E model}
    \label{fig:factor_graph_e}
\end{subfigure}
\hfill
\begin{subfigure}[b]{0.48\columnwidth}
  \centering
  \resizebox{\columnwidth}{!}{
  \tikz {
    \node [circle,draw] (A)              {A};
    \node [circle,draw] (i) [right=0.1cm of A] {$\alpha_i$};
    \node [draw,dotted,fit=(A) (i)] (Ai) {};
    \node [draw] (AiR) [below=0.2cm of Ai,inner sep=0pt,minimum size=5pt] {};
    \node [circle,draw] (R) [below=0.2cm of AiR] {R};
    \node [draw] (ijk) [above right=0.03cm and 0.3cm of R,inner sep=0pt,minimum size=5pt] {};
    \node [circle,draw] (j) [below left=0.5cm and 0.15cm of R] {$\alpha_j$};
    \node [circle,draw] (B) [left=0.2cm of j] {B};
    \node [draw] (RjB) [above=0.6cm of $(B)!0.5!(j)$,inner sep=0pt,minimum size=5pt] {};
    \node [circle,draw] (C) [below right=0.5cm and 0.15cm of R] {C};
    \node [circle,draw] (k) [right=0.2cm of C] {$\alpha_k$};
    \node [draw] (RkC) [above=0.6cm of $(C)!0.5!(k)$,inner sep=0pt,minimum size=5pt] {};
    \draw [->]
    (AiR) edge (R)
    (ijk) edge[bend left=45] (j)
    (ijk) edge[bend left=45] (k)
    (RjB) edge (B)
    (RkC) edge (C)
    (Ai) edge (AiR)
    (i) edge (ijk)
    (R) edge (ijk)
    (R) edge (RjB)
    (j) edge (RjB)
    (R) edge (RkC)
    (k) edge (RkC);
  }}
  \caption{P model}
    \label{fig:factor_graph_p}
\end{subfigure}
  \caption{Extended factor graph notation of decomposed binary rules~\cite{10.5555/2100584.2100615}. Each square \raisebox{-1pt}{\tikz{\node [draw] () {};}} represents a factor. Arrows indicate conditional probabilities. }
  \label{fig:factor_graph}
\end{figure}

Marginalizing $t_2$ in Neural QCFG has a high time complexity of $O(|\NT|(|\NT|+|\PT|)^2S^3T^3)$ where $S/T$ are the source/target sequence lengths. In particular, the number of rules in QCFG contributes to a significant proportion, $O(|\NT|(|\NT|+|\PT|)^2S^3)$, of the complexity. Below, we try to reduce this complexity by rule decompositions in two ways.

\subsection{Efficient Model (E Model)}

Let $\mathcal{R}$ be a new set of symbols. The E model decomposes binary rules $r_b$ into three parts: $A[\alpha_i]\rightarrow R, R\rightarrow B[\alpha_j]$ and $R\rightarrow C[\alpha_k]$ (Fig.~\ref{fig:factor_graph_e}), where $R\in\mathcal{R}$
such that
\begin{align*}
  & p(A[\alpha_i] \rightarrow B[\alpha_j] C[\alpha_k]) = \sum_R p(A[\alpha_i]\rightarrow R) \\
  & \quad  \times p(R\rightarrow B[\alpha_j]) \times p(R\rightarrow C[\alpha_k]).
\end{align*}
In this way, $|\NT|(|\NT|+|\PT|)^2S^3$ binary rules are reduced to only $G_E\coloneqq(3|\NT| + 2|\PT|)|\mathcal{R}|S$ decomposed rules, resulting in a time complexity of $O(G_ET^3)$\footnote{Typically, we set $|R| = O(|\NT|+|\PT|)$.} for marginalizing $t_2$. 
Further, the complexity can be improved to $O(|\mathcal{R}|T^3 + |\mathcal{R}|^2T^2)$ using rank-space dynamic programming in \citet{yang-etal-2022-dynamic}\footnote{They describe the algorithm using TN-PCFG~\citep{yang-etal-2021-pcfgs}, which decomposes binary rules of PCFG, $A\rightarrow BC$, into $A\rightarrow R, R\rightarrow B$ and $R\rightarrow C$. For our case, one can define new symbol sets by coupling nonterminals with source tree nodes: $\mathcal{N}_t = \{(A, \alpha_i)| A\in \NT, \alpha_i\in t_1\}$ and $\mathcal{P}_t = \{(A, \alpha_i)| A\in \PT, \alpha_i\in t_1\}$. Then our decomposition becomes identical to TN-PCFG and their algorithm can be applied directly. }.

However, constraints that simultaneously involve $\alpha_i, \alpha_j,\alpha_k$ (such as the tree hierarchy constraint in vanilla Neural QCFG and those to be discussed in Sec.~\ref{sec:soft_constraint}) can no longer be imposed because of two reasons. First, the three nodes are in separate rules and enforcing such constraints would break the separation and consequently undo the reduction of time complexity. 
Second, the rank-space dynamic programming algorithm prevents us from getting the posterior distribution $p(\alpha_i,\alpha_j,\alpha_k | t_1,s_2)$, which is necessary for many methods of learning with constraints~\cite[e.g.,][]{10.5555/1620270.1620322,10.1145/1273496.1273571,10.5555/1756006.1859918} to work. 

\subsection{Expressive Model (P Model)}

In the P model, we reserve the relation among $\alpha_i,\alpha_j,\alpha_k$ and avoid their separation,
\begin{align*}
  & p(A[\alpha_i] \rightarrow B[\alpha_j] C[\alpha_k]) = \\
  & \quad  \sum_R \begin{aligned}[t]
    & p(A[\alpha_i]\rightarrow R) \times p(R,\alpha_i\rightarrow \alpha_j,\alpha_k) \times \\
    & p(R,\alpha_j\rightarrow B)\times p(R,\alpha_k\rightarrow C),
    \end{aligned}
\end{align*}
as illustrated in Fig.~\ref{fig:factor_graph_p}. 
The P model is still faster than vanilla Neural QCFG because there are only $G_P\coloneqq |\mathcal{R}|S^3 + (3|\NT|+2|\PT|)|\mathcal{R}|S$ decomposed rules, which is lower than vanilla Neural QCFG but higher than the E model. However, unlike the E model, the P model cannot benefit from rank-space dynamic programming\footnote{Below is an intuitive explanation. Assume there is only one nonterminal symbol. Then we can remove $A,B,C$ because they are constants. The decomposition can be simplified to $\alpha_i\rightarrow R, R\alpha_i\rightarrow\alpha_j\alpha_k$, which is equivalent to $\alpha_i\rightarrow\alpha_j\alpha_k$, an undecomposed binary rule. The concept ``rank-space'' is undefined in an undecomposed PCFG.} and has a complexity of  $O(|\mathcal{R}|S^2T^3 + ((2|\NT|+|\PT|)|\mathcal{R}|S + |\mathcal{R}|S^3)T^2)$ for marginalizing $t_2$\footnote{It is better than $O(G_PT^3)$ because we can cache some intermediate steps, as demonstrated in \citet{cohen-etal-2013-approximate,yang-etal-2021-pcfgs}. Details can be found in Appx.~\ref{sec:time_complexity_p_model}.}.

Rule $R,\alpha_i\rightarrow \alpha_j,\alpha_k$ is an interface for designing constraints involving $\alpha_i,\alpha_j,\alpha_k$.
For example, by setting $p(R,\alpha_1\rightarrow \alpha_2,\alpha_3) = 0$ for all $R\in \mathcal{R}$ and certain $\alpha_i,\alpha_j,\alpha_k$, we can prohibit the generation $A[\alpha_1]\rightarrow B[\alpha_2]C[\alpha_3]$ in the original QCFG. With this interface, the P model can impose all constraints used by vanilla Neural QCFG as well as more advanced constraints introduced next section.

\section{Constraints}

\subsection{Soft Tree Hierarchy Constraint}
\label{sec:soft_constraint}

Denote the distance between two tree nodes\footnote{The distance between two tree nodes is the number of edges in the shortest path from one node to another.} as $d(\alpha_i,\alpha_j)$ and define $d(\alpha_i,\alpha_j)=\infty$ if $\alpha_j$ is not a descendant of $\alpha_i$. Then, the distance of a binary rule is defined as $d(r)=\max(d(\alpha_i,\alpha_j), d(\alpha_i,\alpha_k))$. 

Neural QCFG is equipped with two hard hierarchy constraints. For $A[\alpha_i]\rightarrow B[\alpha_j]C[\alpha_k]$,  $\alpha_j, \alpha_k$ are forced to be either descendants of $\alpha_i$ (i.e., $d(r)<\infty$), or more strictly, distinct direct children of $\alpha_i$ (i.e., $d(r)=1$). However, we believe the former constraint is too loose and the latter one is too tight. Instead, we propose a soft constraint based on distances: rules with smaller $d(r)$ are considered more plausible. Specifically, we encode the constraint into a reward function of rules, $\zeta(d(r))$, such that $\zeta(1)>\zeta(2)>\dots$ and
$\zeta(a)\zeta(b) > \zeta(c)\zeta(d)$ for $a+b=c+d$ and $\max(a,b)<\max(c,d)$. A natural choice of the reward function is $\zeta(d(r))\coloneqq d(r)e^{-d(r)}$.
We optimize the expected rewards with a maximum entropy regularizer~\cite{doi:10.1080/09540099108946587,10.5555/3045390.3045594}, formulated as follows:
\begin{align*}
  & \log\sum_{t_2\in\T(s_2)} p_\theta(t_2|t_1) \zeta(t_2) + \tau\mathbb{H}\left(p_\theta(t_2|t_1, s_2)\right),
\end{align*}
where $\zeta(t_2)=\prod_{r\in t_2} \zeta(d(r))$\footnote{$r\in t_2$ means the rule at each generation step of $t_2$.}, $p_\theta(t_2|t_1, s_2) = p_\theta(t_2 | t_1)/\sum_{t\in\T(s_2)} p_\theta(t | t_1)$, $\mathbb{H}$ represents entropy, and $\tau$ is a positive scalar.

\subsection{Coverage Constraint}
\label{sec:coverage_constraint}

Our experiments on vanilla neural QCFG show that inferred alignments could be heavily imbalanced: some source tree nodes are aligned with multiple target tree nodes, while others are never aligned. 
This motivates us to limit the number of alignments per source tree node with an upper bound\footnote{We do not set lower bounds, meaning each source tree node should be aligned at least $n$ times, because our source-side parser uses a grammar in CNF, and such trees could contain semantically meaningless nodes, which are not worthing to be aligned. For example, trees of \textit{Harry James Potter} must contain either \textit{Harry James} or \textit{James Potter}.}, $u$. 
Because the total number of alignments is fixed to $|t_2|$, this would distribute alignments from popular source tree nodes to unpopular ones, leading to more balanced source coverage of alignments.
We impose this constraint via optimizing the posterior regularization likelihood~\cite{10.5555/1756006.1859918},
\begin{equation*}
\resizebox{\hsize}{!}{
$\mathbb{E}_{t_1} \left(\log p_\theta(s_2|t_1) + \gamma\min_{q\in \mathcal{Q}} \mathbb{KL}(q(t_2) || p_\theta(t_2|t_1,s_2))\right),$
}
\end{equation*}
where $\mathbb{KL}$ is the Kullback-Leibler divergence (KL), $\gamma$ is a positive scalar and $\mathcal{Q}$ is the constraint set $\{q(t_2)|\mathbb{E}_{q(t)} \phi(t) \le \xi\}$, i.e., expectation of feature vector $\phi$ over any distribution in $\mathcal{Q}$ is bounded by constant vector $\xi$. 
We define the target tree feature vector $\phi(t_2)\in \mathbb{N}^{|t_1|}$ such that $\phi_i(t_2)$ represents the count of source tree node $\alpha_i$ being aligned by nodes in $t_2$ and $\xi=u\bm{1}$.
\citet{10.5555/1756006.1859918} provide an efficient algorithm for finding the optimum $q$, which we briefly review in Appx.~\ref{sec:more_coverage_constraint}. After finding $q$, the KL term of two tree distributions, $q$ and $p_\theta$, can be efficiently computed using the \sffamily Torch-Struct \rmfamily library~\cite{rush-2020-torch}.

\section{Experiments}

We conduct experiments on the three datasets used in \citet{NEURIPS2021_dd17e652}.
Details can be found in Appx.~\ref{sec:exp_detail}.

\subsection{SCAN}

\begin{table}[tb]
\centering
\bgroup
\setlength\tabcolsep{5pt}
\begin{tabular}{ccccc} \toprule
Approach & Simple & Jump & A. Right & Length \\\midrule
vNQ$^1$ & 96.9 & 96.8 & 98.7 & 95.7 \\
E\textsubscript{Model} & 9.01 & - & 1.2 & -  \\
P\textsubscript{Model} & 95.27 & 97.08 & 97.63 & 91.72 \\\bottomrule  
\end{tabular}
\egroup
\caption{Accuracy on the SCAN datasets. vNQ$^1$ is vanilla Neural QCFG from \citet{NEURIPS2021_dd17e652}. vNQ$^1$ and P\textsubscript{Model} use the hard constraint $d(r)<\infty$.}
\label{tab:scan}
\end{table}

We first evaluate our models on four splits of the SCAN dataset~\cite{lake2018generalization}. We report accuracy in Tab.~\ref{tab:scan}. The P model equipped with constraints can achieve almost perfect performance similar to vanilla Neural QCFG, while the E model fails due to a lack of constraints.

\subsection{Style Transfer and En-Fr Translation}

\newcommand{\uf}{\textcolor{red}{00.00}}
\begin{table}[tb]
\centering
\bgroup
\setlength\tabcolsep{4pt}
\begin{tabular}{cccccc} \toprule
Approach & \textit{nil} & +H$^1$ & +H$^2$ & +S & +C \\\midrule
\multicolumn{6}{l}{\cellcolor{gray!10}\textit{Active to passive (ATP)}} \\
vNQ$^1$ & $-$ & 66.2 & $-$ & $-$ & $-$ \\
vNQ$^2$ & 71.42 & 71.56 & $-$ & 71.62 & 73.86 \\
E\textsubscript{Model}& 73.48 & $\times$ & $\times$ & $\times$ & 74.25 \\
P\textsubscript{Model}& 75.06 & 69.88 & $-$ & 73.11 &  75.44  \\
\multicolumn{6}{l}{\cellcolor{gray!10}\textit{Adjective Emphasise (AEM)}} \\
vNQ$^1$ & $-$ & 31.6 & $-$ & $-$ & $-$ \\
vNQ$^2$ & 28.82 & 31.52 & $-$ & 36.77 & 30.81 \\
E\textsubscript{Model}& 28.33 & $\times$ & $\times$ & $\times$ & 28.67 \\
P\textsubscript{Model}& 31.81 & 29.14& $-$ & 35.91  & 30.12  \\
\multicolumn{6}{l}{\cellcolor{gray!10}\textit{Verb Emphasise (VEM)}} \\
vNQ$^1$ & $-$ & 31.9 & $-$ & $-$ & $-$ \\
vNQ$^2$ & 26.09 & 29.64 & $-$ & 30.50 & 28.50 \\
E\textsubscript{Model}& 25.21 & $\times$ & $\times$ & $\times$ & 24.67 \\
P\textsubscript{Model}& 27.43 & 24.77 & $-$ & 26.81 & 30.66 \\
\multicolumn{6}{l}{\cellcolor{gray!10}\textit{En-Fr machine translation}} \\
vNQ$^1$ & $-$ & $-$  & 26.8 & $-$ & $-$ \\
vNQ$^2$ & 28.63 & $-$ & 29.10 & 30.45 & 31.87 \\
E\textsubscript{Model}& 28.93 & $\times$ & $\times$ & $\times$ & 29.33 \\
P\textsubscript{Model}& 29.27 & $-$ & 29.76& 30.51 & 29.69 \\\bottomrule
\end{tabular}
\egroup
\caption{BLEU-4 for tasks from the StylePTB dataset (the top three series) and BLEU for Fr-En machine translation against different models and constraints. vNQ$^2$ is our reimplementation of \citet{NEURIPS2021_dd17e652}. \textit{nil} means that no constraint is placed. H$^1$ and H$^2$ is the hard constraint $d(r)<\infty$ and $d(r)=1$, respectively. S is the soft tree hierarchy constraint. C is the coverage constraint. $\times$ means that the constraint is inapplicable and $-$ means we do not run the experiment or \citet{NEURIPS2021_dd17e652} does not report the score.}
\label{tab:styleptb_constraint}
\end{table}

\begin{figure}[tb]
    \centering
    \includegraphics[width=\columnwidth]{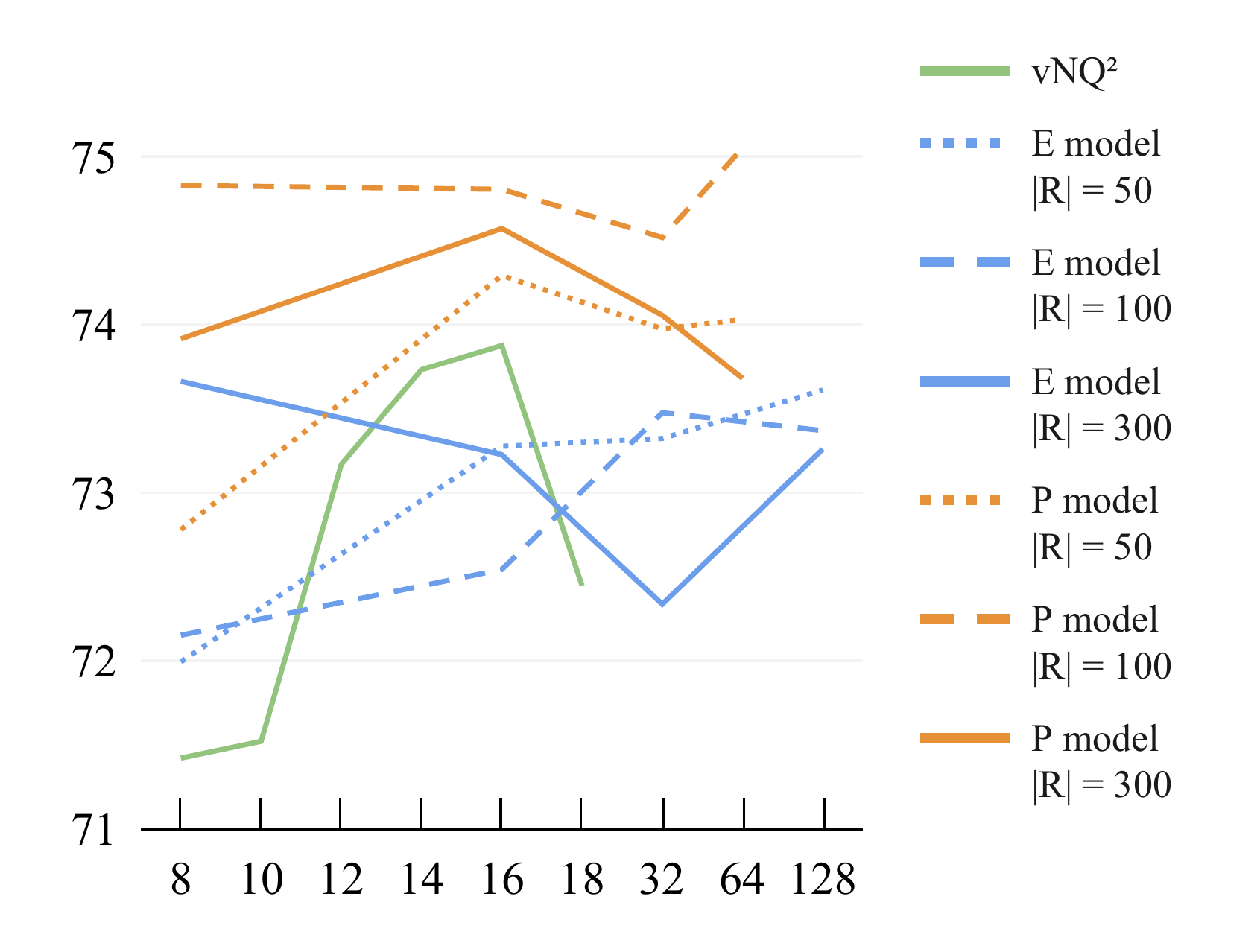}
    \caption{BLEU-4 scores on the ATP task. No constraint is placed. The horizontal axis represents $|\NT|(= |\PT|)$. }
    \label{fig:styleptb_size}
\end{figure}

Next, we evaluate the models on the three hard transfer tasks from the StylePTB dataset~\cite{lyu-etal-2021-styleptb} and a small-scale En-Fr machine translation dataset~\cite{lake2018generalization}.
Tab.~\ref{tab:styleptb_constraint} shows results of the models with different constraints\footnote{Following \citet{NEURIPS2021_dd17e652}, we calculate the metrics for tasks from the StylePTB dataset using the \sffamily nlg-eval \rmfamily library (\citet{sharma2017nlgeval}; \url{https://github.com/Maluuba/nlg-eval}) and calculate BLEU for En-Fr MT using the multi-bleu script (\citet{koehn-etal-2007-moses}; \url{https://github.com/moses-smt/mosesdecoder}). 
}. 
Low-rank models generally achieve comparable or better performance and consume much less memory\footnote{We report speed and memory usage briefly in Sec~\ref{sec:speed_summary} and in detail in Appx.~\ref{sec:speed_and_memory}.}.
We can also find that the soft tree hierarchy constraint outperforms hard constraints and is very helpful when it comes to extremely small data (i.e., AEM and VEM). The coverage constraint also improves performance in most cases.

\subsection{Analysis}
We study how the number of nonterminals affects performance. On our computer\footnote{One NVIDIA TITIAN RTX with 24 GB memory.}, we can use at most 18/64/128 nonterminals in vanilla Neural QCFG/the P model/the E model, showing that our low-rank models are more memory-friendly than vanilla Neural QCFG.
We report results in Fig.~\ref{fig:styleptb_size}. 
There is an overall trend of improved performance with more nonterminals (with some notable exceptions).
When the numbers of nonterminals are the same, the P model outperforms vanilla Neural QCFG consistently, showing its superior parameter efficiency. In contrast, the E model is defeated by vanilla QCFG and the P model in many cases, showing the potential harm of separating $\alpha_i,\alpha_j,\alpha_k$. 

\subsection{Speed Comparison}
\label{sec:speed_summary}

\begin{figure}[tb]
    \centering
    \includegraphics[width=\columnwidth]{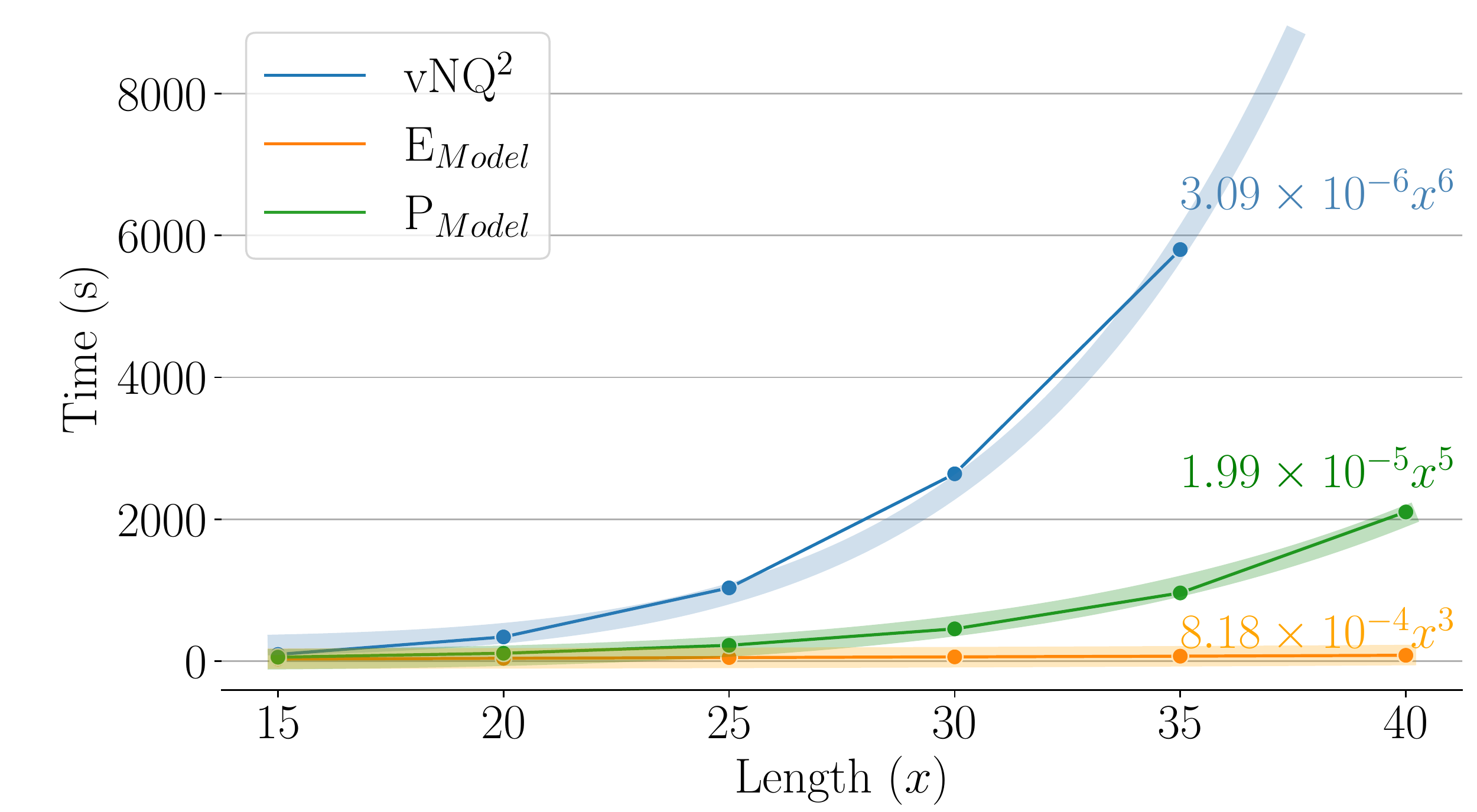}
    \caption{The duration required to train one epoch on synthetic datasets with different length ($x=S=T$). Thick and shallow lines are fitted curves based on time complexities of vNQ$^2$, E$_{Model}$ and P$_{Model}$, i.e., $O(x^6), O(x^3)$ and $O(x^5)$.}
    \label{fig:length_time}
\end{figure}

\begin{figure}[tb]
    \centering
    \includegraphics[width=\columnwidth]{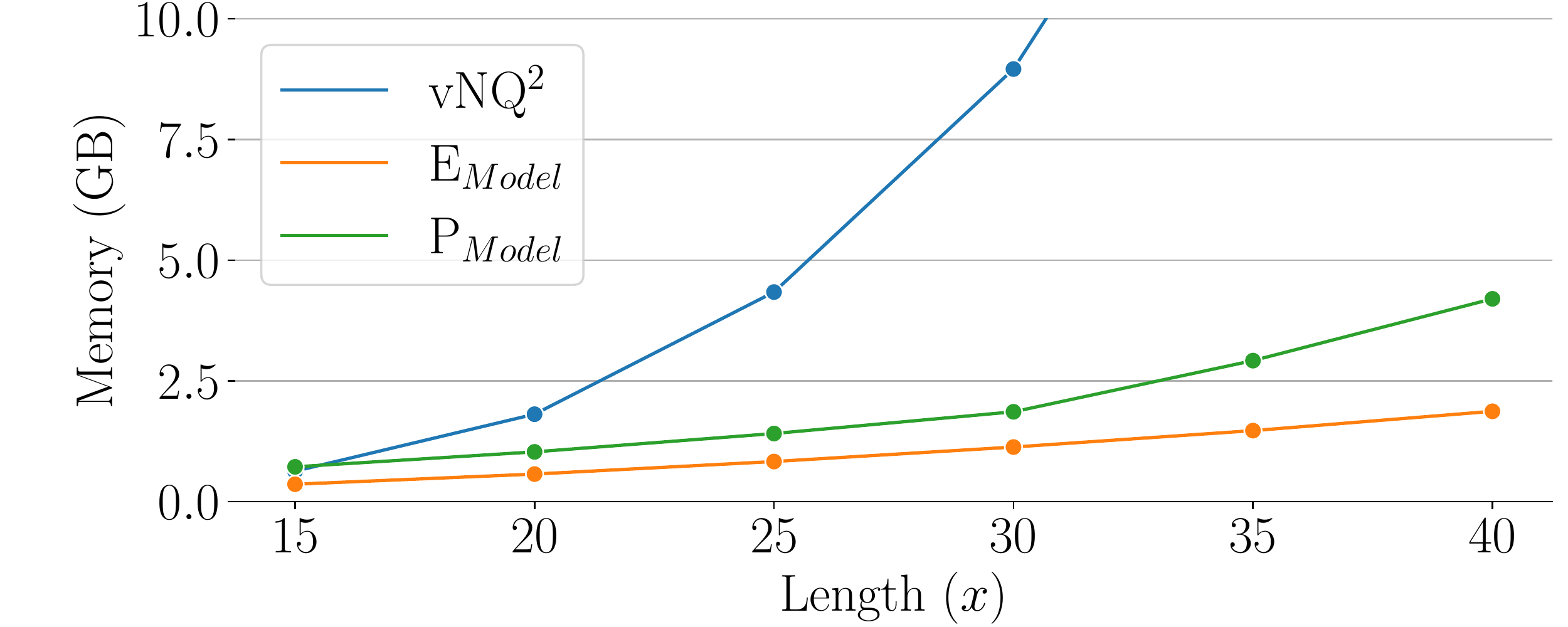}
    \caption{Memory usage for training with batch size 1 on synthetic datasets with different length ($x=S=T$).}
    \label{fig:length_memory}
\end{figure}

We benchmark speed and memory usage using synthetic datasets with different sequence lengths. Fig.~\ref{fig:length_time} and \ref{fig:length_memory} illustrate the results. Compared to the standard Neural QCFG, the E model and P model are significantly faster and have a lower memory footprint. This enables them to model longer sequences effectively. For data construction and more results, please refer to Appx.~\ref{sec:speed_and_memory}.

\section{Conclusion}

We have presented two low-rank variants of Neural QCFG based on decomposition for efficiency and two new constraints over tree hierarchy and source coverage. 
Experiments on three datasets validate the effectiveness and efficiency of our proposed models and constraints. 

\section{Limitations}

First, unlike decoders in neural seq2seq models, which can attend to any previously generated tokens, QCFGs have a strong context-free independence assumption during generation. With this assumption, Neural QCFG cannot model some complex distributions. A potential solution is to use stronger grammars, such as RNNG~\cite{dyer-etal-2016-recurrent} and Transformer Grammars~\cite[TG;][]{10.1162/tacl_a_00526}.

Second, we assume that both the grammars used by the source-side parser and QCFG are in CNF. Although it is convenient for discussion and implementation, CNF does not suit for modeling the structure of practical sequences. In semantic representations (e.g., Abstract Meaning Representation~\cite{banarescu-etal-2013-abstract}), a predicate could have more than two arguments. Ideally, we should represent $n$-ary predicates with $n$-ary rules. However, for grammars in CNF, $n-1$ unnatural binary rules are required to represent $n$-ary predicates. In natural language, we will face semantically meaningless spans due to CNF, which is discussed in Sec~\ref{sec:coverage_constraint}.

Third, although using decomposition improves the speed and the memory requirement, our low-rank models still cost much more computation resources than neural \ac{seq2seq} models for two main reasons. (1) A large amount of nonterminal symbols increase the memory cost significantly. (2) Because finding the most probable string $t_2$ from $p_\theta(t_2 |t_1)$ is NP-hard~\cite{simaan-1996-computational,LYNGSO2002545}, we follow \citet{NEURIPS2021_dd17e652} to use a decoding strategy with heavy sampling. For real data, we may need to sample hundreds or thousands of sequences and then rank them, which can be much slower than the decoding of neural seq2seq models.

\section*{Acknowledgments}
We thank the anonymous reviewers for their constructive comments. This work was supported by the National Natural Science Foundation of China (61976139).


\bibliography{anthology,custom}
\bibliographystyle{acl_natbib}

\appendix

\section{Time Complexity of P Model}
\label{sec:time_complexity_p_model}

  
Let $\beta_{ij}, \beta_{jk} \in \mathbb{R}^{|\NT|\times|t_1|}$ be two cells in the chart of the dynamic programming. $\beta_{ij}(x,y)$ denotes indexing into the matrix. Denote $A[\alpha_1] \rightarrow B[\alpha_2] C[\alpha_3]$ as $r_b$. 
The state transition equation is 
\begin{align*}
 \beta_{ik}(A,\alpha_1) = \sum_{\substack{j,B,C\\ \alpha_2,\alpha_3}} p(r_b)\beta_{ij}(B,\alpha_2)\beta_{jk}(C,\alpha_3).
\end{align*}
Let's define following terms:
\begin{align*}
    &\tilde{\beta}_{ij}(R,\alpha_2) = \sum_B p(R,\alpha_2\rightarrow B) \beta_{ij}(B,\alpha_2) \\
    &\tilde{\beta}_{jk}(R,\alpha_3) = \sum_C p(R,\alpha_3\rightarrow C) \beta_{ij}(C,\alpha_3) \\
    & \hat{p} = p(A[\alpha_1]\rightarrow R) p(R,\alpha_1\rightarrow \alpha_2,\alpha_3)
\end{align*}
Then the state transition equation can be reformulated as:
\begin{align*}
  \beta_{ik}(A,\alpha_1) = \sum_{R,\alpha_2,\alpha_3} \hat{p} \underbrace{\sum_j \tilde{\beta}_{ij}(R,\alpha_2) \tilde{\beta}_{jk}(R,\alpha_3)}_{\hat{\beta}_{ik}},
\end{align*}
where $\hat{\beta}_{ij}\in\mathbb{R}^{|\mathcal{R}|\times |t_1| \times |t_1|}$. We can compute $\tilde{\beta}_{ij}$ in $O((|\NT|+|\PT|)|\mathcal{R}|S)$ and cache it for composing $\hat{\beta}_{ij}$. Then $\hat{\beta}_{ik}$ can be computed in $O(|\mathcal{R}|S^2T)$. Finally, we can compute $\beta_{ik}$ in $O(|\mathcal{R}|S^3 + |\NT||\mathcal{R}|S)$ by sum out $\alpha_2,\alpha_3$ first:
\begin{align*}
    &\beta_{ik}(A,\alpha_1) = \\
    &\sum_R p(A[\alpha_1]\rightarrow R) \sum_{\alpha_2,\alpha_3} p(R,\alpha_1\rightarrow \alpha_2,\alpha_3) \hat{\beta}_{ik}
\end{align*}

So, summing terms of all the above steps and counting the iteration over $i,k$, we will get $O(|\mathcal{R}|S^2T^3 + ((2|\NT|+|\PT|)|\mathcal{R}|S + |\mathcal{R}|S^3)T^2)$.

\section{Neural Parameterization}
\label{sec:parameterization}

We mainly follow \cite{NEURIPS2021_dd17e652} to parameterize the new decomposed rules. First, we add embeddings of terms on the same side together. For example, we do two additions $e_{lhs} = e_{R}+e_{\alpha_i}$ and $e_{rhs}=e_{\alpha_j}+e_{\alpha_k}$ for $R,\alpha_i\rightarrow \alpha_j,\alpha_k$, where $e_x$ denotes the embedding of $x$. Note that we use the same feed-forward layer $f$ as \cite{NEURIPS2021_dd17e652} to obtain $e_x$ from some feature $h_x$. i.e. $e_x=f(h_x)$. Then, we compute the inner products of embeddings obtained in the previous step as unnormalized scores. For example, $p(R,\alpha_i\rightarrow \alpha_j,\alpha_k)\propto \exp(e_{lhs}^\top e_{rhs})$.

\section{Posterior Regularization}
\label{sec:more_coverage_constraint}

The problem $\min_{q\in\mathcal{Q}} \mathbb{KL}(q(t_2)||p(t_2|t_1,s_2))$ has the optimal solution $$
  q^* = \frac{1}{Z(\lambda^*)} p(t_2|t_1,s_2) \exp\{-\lambda^* \phi(t_2)\},
$$
where $$Z(\lambda^*)=\sum_{t_2} p(t_2|s_1,t_1)\exp\{-\lambda^* \phi(t_2)\}$$ and $\lambda^*$ is the solution of the dual problem:
$$
  \max_{\lambda \ge 0} -b \cdot \lambda - \log Z(\lambda)
$$

We can reuse the inside algorithm to compute $Z(\lambda^*)$ efficiently because our $\phi(t)$ can be factored as $p(t_2|t_1,s_2)$:
\begin{align*}
  p(t_2|t_1, s_2) &=  \prod_{r\in t_2} p_\theta(r) \\
  \phi(t) &= \sum_{r\in t_2} \phi(r, t_1),
\end{align*}
where $\phi(r,t_1) = 1$ if $t_1$ is in the left-hand side of $r$ and $\phi(r,t_1) = 0$ otherwise. Then, the solution $q^*$ can be written as $$
  q^*(t_2) \propto \prod_{r\in t_2} p_\theta(r) \exp\{-\lambda \phi(r, t_1)\}.
$$

Recall that we define $\phi(t)$ to be the counts of source nodes being aligned by nodes in $t$. We can factor $\phi(t)$ in terms of $r$ because each target tree non-leaf node invokes exactly one rule and only occurs on the left-hand side of that rule. So, the sum over $r$ is equivalent to the sum over target tree nodes. 

\section{Experiments}

\subsection{Experimental Details}
\label{sec:exp_detail}

We implement vNQ$^2$, the E model, and the P model using our own codebase. We inherit almost all hyperparameters of \citet{NEURIPS2021_dd17e652} and a basic constraint: the target tree leaves/non-leaf nodes can only be aligned to source tree leaves/non-leaf nodes, and especially, the target tree root can only be aligned to the source tree root. One major difference is that, in our experiments, we do not use early-stopping and run fixed optimization steps, which are much more than the value set in \citet{NEURIPS2021_dd17e652} (i.e., 15). It is because in preliminary experiments\footnote{We run 100 epochs and evaluate task metrics on validation sets every 5 epochs.}, we found that the task metric (e.g., BLEU) almost always get improved consistently with the process of training, while the lowest perplexity occurs typically at an early stage (which is the criteria of early-stopping in \citet{NEURIPS2021_dd17e652}), and computing task metric is very expensive for Neural QCFGs.  We report metrics on test sets averaged over three runs on all datasets except for SCAN. As mentioned in the code of \citet{NEURIPS2021_dd17e652}, we need to run several times to achieve good performance on SCAN. Therefore, we report the maximum accuracy in twenty runs. 

\noindent\textbf{SCAN} \cite{lake2018generalization} is a diagnostic dataset containing translations from English commands to machine actions. We conduct experiments on four splits: 
We evaluate our models on four splits of the SCAN~\cite{lake2018generalization} dataset: \textit{simple}, \textit{add primitive (jump)}, \textit{add template (around right)} and \textit{length}. The latter three splits are designed for evaluating compositional generalization. Following \cite{NEURIPS2021_dd17e652}, we set $|\NT|=10, |\PT|=1$.

\noindent\textbf{StylePTB} \cite{lyu-etal-2021-styleptb} is a text style tranfer dataset built based on Penn Treebank~\cite[PTB;][]{marcus-etal-1993-building}. Following \citet{NEURIPS2021_dd17e652}, we conduct experiments on three hard transfer tasks: textit{active to passive} (2808 examples), \textit{adjective emphasis} (696 examples) and \textit{verb emphasis} (1201 examples). According to Tab.~\ref{fig:styleptb_size}, we set $|\NT|=|\PT|=32,|\mathcal{R}|=100$ for the E model and set $|\NT|=|\PT|=64,|\mathcal{R}|=100$ for the P model. 

\noindent\textbf{En-Fr MT} \cite{lake2018generalization} is a small-scale machine translation dataset. We use the same split as \citet{NEURIPS2021_dd17e652}. The size of training/validate/test set is 6073/631/583. We set $|\NT|=|\PT|=32,|\mathcal{R}|=100$ for the E model and $|\NT|=|\PT|=32,|\mathcal{R}|=196$ for the P model.

\subsection{Tune Hyperparameter}

We tune hyperparameters according to metrics on validation sets, either manually or with the Bayesian Optimization and Hyperband (BOHB) search algorithm~\cite{Falkner2018BOHBRA} built in the \sffamily wandb \rmfamily library. First, we tune $|\NT|$, $|\PT|$, $|\mathcal{R}|$ and the learning rate of parameters for parameterizing QCFG. We freeze hyperparameters related to the source-side parser, the contextual encoder (i.e., LSTM), and the TreeLSTM~\cite{tai-etal-2015-improved,pmlr-v37-zhub15}. For the ATP task from StylePTB, we run the grid search to plot Fig.~\ref{fig:styleptb_size} and choose the best hyperparameters. For other tasks, we run about 20 trials according to BOHB for each manually set search range. Typically, the size of a search range is 256 (four choices for each tunable hyperparameter). Next, we tune the strength of the coverage constraint for all models by running with $\gamma=0.5, 1, 2$.

\subsection{Speed and Memory Usage Comparison}
\label{sec:speed_and_memory}

Tab.~\ref{tab:speed} shows the time and memory usage on synthetic datasets. Each synthetic dataset contains 1000 pairs of random sequences with the same length sampled from a vocabulary with size 5000, i.e., $\{(s_1, s_2)_1, \dots (s_1, s_2)_{1000}\}, s_1, s_2 \in \Sigma^{v}, |\Sigma|=5000$ where $v$ is the length.
We set $|\NT|=|\PT|=8$ for vanilla Neural QCFG and $|\NT|=|\NT|=50, |\mathcal{R}|=200$ for others. We train models on a computer with an NVIDIA GeForce RTX3090. Note that we disable the copy mechanism in \citet{NEURIPS2021_dd17e652} because of its complicated effects on memory usage, such that the results differ from Fig.~\ref{fig:styleptb_size} (in which models enable the copy mechanism).

\begin{table*}[t]
    \centering
\begin{tabular}{cccccc}\toprule
\multicolumn{1}{l}{$v$} & Approach              & Constraint & Batch size           & Time (s)             & GPU Memory (GB)      \\\midrule
\multirow{12}{*}{10}    & \multirow{5}{*}{vNQ$^2$} & \textit{nil} & 8 & 25.6 & 1.42 \\
 & & +H$^1$ & 8 & 25.5 & 1.43 \\
 & & +H$^2$ & 8 & 113.8 & 7.67\\
 & & +S & 8 & 60.5 & 2.46 \\
 & & +C & 8 & 132.7 & 3.08 \\\cline{2-6}
 & \multirow{2}{*}{E\textsubscript{Model}} & \textit{nil} & 8 & 20.1 & 1.59 \\
 & & +C & 8 & 40.4 & 1.59 \\\cline{2-6}
 & \multirow{5}{*}{P\textsubscript{Model}} & \textit{nil} & 8 & 30.7 & 3.78 \\
 & & +H$^1$ & 8 & 31.3 & 3.79 \\
 & & +H$^2$ & 8 & 64.0 & 6.41 \\
 & & +S & 8 & 45.8 & 4.08 \\
 & & +C & 8 & 73.9 & 4.02 \\\midrule
\multirow{12}{*}{20}    & \multirow{5}{*}{vNQ$^2$} & \textit{nil} & 8 & 341.2 & 14.49 \\
 & & +H$^1$ & 8 & 342.4 & 14.60 \\
 & & +H$^2$ & 1 & $\approx$16539.4 & 14.13 \\
 & & +S & 2 & $\approx$1734.4 & 8.93 \\
 & & +C & 2 & $\approx$4657.1 & 12.24 \\\cline{2-6}
 & \multirow{2}{*}{E\textsubscript{Model}} & \textit{nil} & 8 & 40.0 & 4.58 \\
 & & +C & 4 & 173.4 & 14.48 \\\cline{2-6}
 & \multirow{5}{*}{P\textsubscript{Model}} & \textit{nil} & 8 & 111.3 & 8.25 \\
 & & +H$^1$ & 8 & 110.8 & 8.29 \\
 & & +H$^2$ & 4 & 452.3 & 9.83 \\
 & & +S & 8 & 269.8 & 18.76 \\
 & & +C & 4 & 643.5 & 18.20 \\\midrule
 \multirow{8}{*}{40}    & \multirow{1}{*}{vNQ$^2$} &  & 1 & $\times$ & $\times$ \\\cline{2-6}
 & \multirow{2}{*}{E\textsubscript{Model}} & \textit{nil} & 8 & 82.5 & 14.95 \\
 & & +C & 8 & 177.0 & 14.95 \\\cline{2-6}
 & \multirow{5}{*}{P\textsubscript{Model}} & \textit{nil} & 4 & $\approx$2102.7 & 16.78 \\
 & & +H$^1$ & 4 & $\approx$2097.6 & 16.96 \\
 & & +H$^2$ & 1 & $\times$ & $\times$ \\
 & & +S & 2 & $\approx$2729.3 & 10.63 \\
 & & +C & 1 & $\times$ & $\times$ \\\bottomrule
\end{tabular}
    \caption{Time and memory usage on synthetic datasets. We report statistics with as large as possbile batch size (in $1,2,4,8$). $\times$ represents that we get an out-of-memory error even if we set batch size to 1. $\approx$ represents that the value is estimated using a small portion of the dataset. }
    \label{tab:speed}
\end{table*}

\end{document}